\documentclass[conference]{IEEEtran}
\IEEEoverridecommandlockouts
\usepackage{cite}
\usepackage{amsmath,amssymb,amsfonts}
\usepackage{algorithmic}
\usepackage{graphicx}
\usepackage{textcomp}
\usepackage{xcolor}
\usepackage{url}

\def\BibTeX{{\rm B\kern-.05em{\sc i\kern-.025em b}\kern-.08em
    T\kern-.1667em\lower.7ex\hbox{E}\kern-.125emX}}
\begin{document}

\title{Road Damages Detection and Classification with YOLOv7
\thanks{Authors will list the funding agency here at the final version.}
}

\author{\IEEEauthorblockN{1\textsuperscript{st} Vung Pham}
\IEEEauthorblockA{\textit{Computer Science Department} \\
\textit{Sam Houston State University}\\
Huntsville, Texas, United States \\
vung.pham@shsu.edu}
\and
\IEEEauthorblockN{2\textsuperscript{nd} Du Nguyen}
\IEEEauthorblockA{\textit{Computer Science Department} \\
\textit{Sam Houston State University}\\
Huntsville, Texas, United States \\
dhn005@shsu.edu}
\and
\IEEEauthorblockN{3\textsuperscript{rd} Christopher Donan}
\IEEEauthorblockA{\textit{Computer Science Department} \\
\textit{Sam Houston State University}\\
Huntsville, Texas, United States \\
cmd106@SHSU.EDU}
}

\maketitle

\begin{abstract}
% road damages
% proposed approach
% experiments
% results
Maintaining the roadway infrastructure is one of the essential factors in enabling a safe, economic, and sustainable transportation system. Manual roadway damage data collection is laborious and unsafe for humans to perform. This area is poised to benefit from the rapid advance and diffusion of artificial intelligence technologies. Specifically, deep learning advancements enable the detection of road damages automatically from the collected road images. This work proposes to collect and label road damage data using Google Street View and use YOLOv7 (You Only Look Once version 7) together with coordinate attention and related accuracy fine-tuning techniques such as label smoothing and ensemble method to train deep learning models for automatic road damage detection and classification. The proposed approaches are applied to the Crowdsensing-based Road Damage Detection Challenge (CRDDC2022), IEEE BigData 2022. The results show that the data collection from Google Street View is efficient, and the proposed deep learning approach results in F1 scores of 81.7\% on the road damage data collected from the United States using Google Street View and 74.1\% on all test images of this dataset.
\end{abstract}

\begin{IEEEkeywords}
road damage, detection, classification, YOLOv7, coordinate attention
\end{IEEEkeywords}

% \section{Requirements}
% \textbf{Contents in technical paper/report (Required):}
% \begin{itemize}
%     \item Explanation of your method, with complete details of technique used (for instance, ensemble learning or data augmentation etc.)
%     \item Evaluation of your method (you can use results obtained on the site of road damage detection challenge to compare)
%     \item Detailed evaluation of your results (inference speed, model size for final deployment, etc.)
% Code and trained model link
% \end{itemize}

% \textbf{Contents in technical paper (Optional but preferable):}
% \begin{itemize}
%     \item Error Analysis: Examples of failed attempts, efforts that did not go well.
%     \item Advantages and limitations of using your proposed solution.

% \end{itemize}

\section{Introduction}
The roadway network is the backbone of the economy. Economic development largely depends on the efficiency, reliability, and safety of highways and transportation systems, which support mobility needs, commerce, and industry. But the roadways face challenges from population growth, deteriorating infrastructure, and rapidly rising roadway construction costs. Maintaining the state's economic momentum requires significant improvements in transportation infrastructure. 

Roadway damages cost U.S. drivers billions of dollars annually. Even worse, they impact the middle- and lower-income individuals more and disproportionately. For instance, pothole damage alone costs U.S. drivers \$3 billion annually~\cite{aaa2016}. Roadway management often needs to collect condition data annually for both pavements and bridges. This data set is critical to maintenance planning efforts and required by other federal reporting needs. The collection of this data set has recently slowed due to budgetary constraints leading to low accuracy and consistency~\cite{mopar2020}.

Therefore, the approach in this proposed work allows us to perform roadway damage data collection efficiently. The proposed approach is expected to provide more data at a lower cost, higher frequency, and better accuracy compared to the manual, unsafe approach. Faster data collection means better responses to roadway issues, reducing maintenance and other economic and opportunity costs, and enhancing citizen safety. 

Additionally, recent advancements in the deep learning area are poised to benefit this area. Therefore, this work explores YOLOv7 and related techniques, such as coordinate attentions and label smoothing, to tackle road damage detection and classification tasks. This work then also apply the selected approaches to the Crowdsensing-based Road Damage Detection Challenge (CRDDC2022), A Track in the IEEE Big Data 2022 Big Data Cup Challenge dataset\footnote{\url{https://rdd2020.sekilab.global/}}. Thus, our contributions are:
\begin{itemize}
    \item Proposing an approach to efficiently collect and label road damages and contributing a road damage dataset to the Crowdsensing-based Road Damage Detection Challenge (CRDDC2022).
    
    \item Exploring current state-of-the-art object detection methods and related techniques to road damage detection and classification tasks.

    \item Experimenting with these approaches using CRDDC2022 dataset.
    
\end{itemize}

\section{Related Work}
Deep learning methods are gaining traction thanks to their state-of-the-art results in various domains such as computer visions~\cite{pham2020scagcnn}, soil science~\cite{pham2021soil, pham2021rdnet}, or even solar flare predictions~\cite{pham2019solar}, naming but a few. The transportation industry is no exception, and the roadway damage identification task is poised to benefit from the rapid advance and diffusion of deep learning technologies. The common deep learning methods in this area include Faster Region-Based Convolutional Neural Networks (Faster R-CNN)~\cite{ren2015faster}, You Only Look Once (YOLO)~\cite{bochkovskiy2020yolov4}, and Single Shot Detection (SSD)~\cite{liu2016ssd}. 

Specific to road damage detection, Arya et al.~\cite{arya2020global} reported a set of state-of-the-art solutions in global roadway damage detection and classification tasks. For instance, Pham et al.~\cite{pham2020road} experimented with Detectron2's implementation of Faster R-CNN implementation in this study. Similarly, Hegde et al.~\cite{hegde2020yet} experimented with YOLO and ensemble approaches on this task. Generally, these reviewed studies show that the Faster R-CNN model provides better accuracy with the trade-off of prediction time ($\approx8$ frames per second) than the YOLO model ($\approx40$ frames per second). Conversely, SSD is the balance between the two regarding prediction accuracy and prediction time\cite{pham2020road}. 

This field is developing rapidly, and there more experiments should be conducted to find a better approach in this specific case. Specifically, among these techniques, YOLO seems to attract more research work because there are more recent releases of this technology. By the time of writing, YOLOv7 was the current YOLO version, and this is what was experimented with extensively in this work. YOLOv7 is the official implementation of a paper~\cite{wang2022yolov7} from Wang et al. 

YOLOv7 outperforms popular object detectors, by the time of writing, in speed and accuracy in the range from five FPS (frames per second) to 160 FPS. It also provides a set of freebies ready to be used and easily finetunes detection models. Furthermore, it is relatively easy to add components/modules to YOLOv7 and create new models using its configuration file. Therefore, this project tested various hyperparameters (using the freebies) and models (by adding custom modules and configuration files) to train models for road damage detection and classification tasks.

\begin{figure}[!htb]
    \centering
    \includegraphics[width=\linewidth]{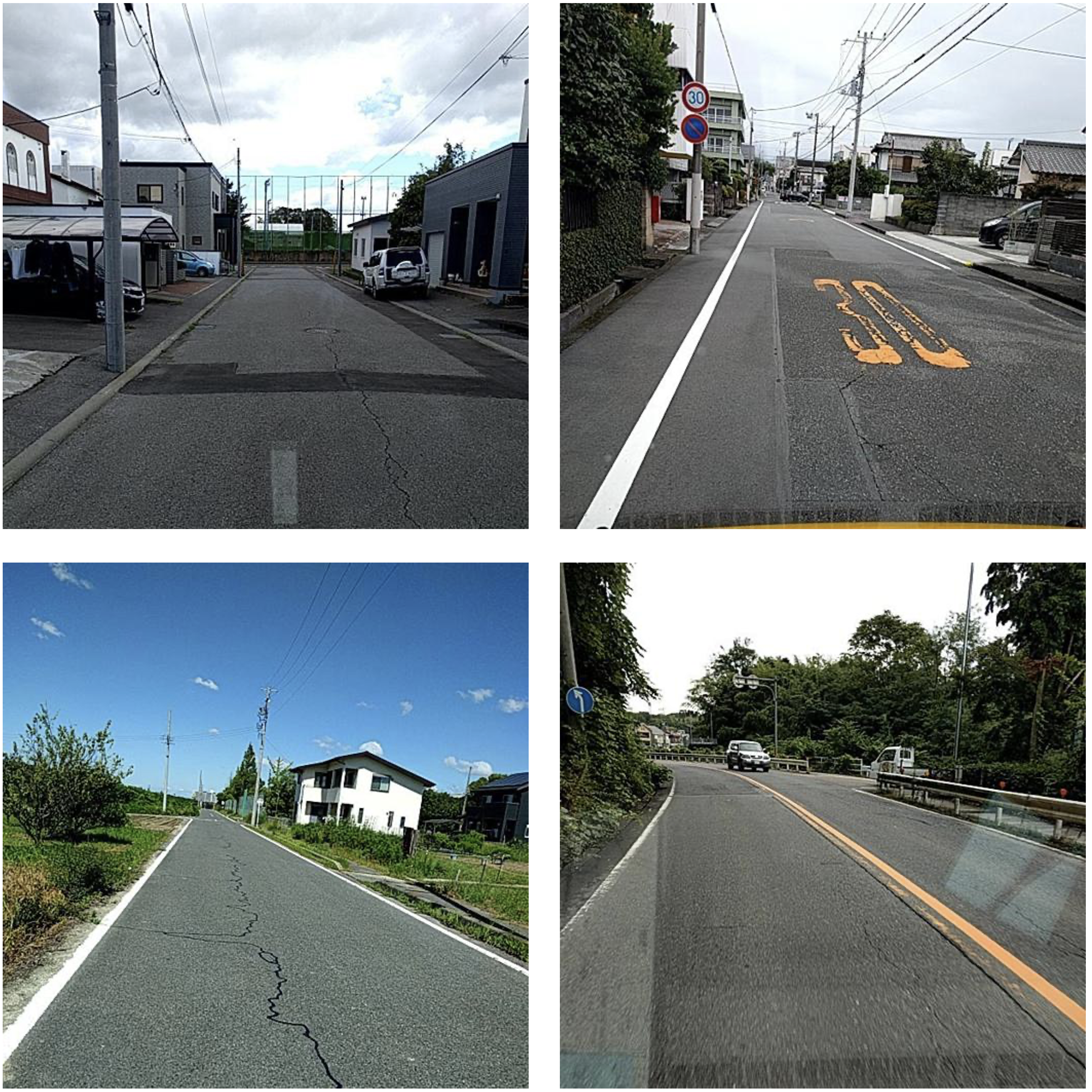}
    \caption{Typical examples of road images collected using cameras placed on car dashboards. These images have typical perspectives. Road damages are more at the lower half of the images, and road damages may look different when they are at different positions on the images.}
    \label{fig:perspectives}
\end{figure}

YOLOv7 back-end networks often make good use of channel attention to weight channels differently while combining channels at different layers. However, channel attention squeezes the whole spatial space of one channel into one value. Thus, it does not capture spatial information. The spatial information is important in road damage detection tasks. Specifically, Figure~\ref{fig:perspectives} shows a typical example of road images collected using a camera (e.g., smartphone camera) placed on a car dashboard (which is a typical road image data collection method) from the RDD2022 (road damage dataset 2022)~\cite{rdd2022}. Observably, most of the road damages are in the lower half of the pictures, and damages at different locations may have different looks due to camera perspectives. Therefore, this work experiments with Coordinate Attention~\cite{hou2021coordinate}. 

The coordinate attention technique adds spatial (coordinate) information into channel attention. Specifically, this technique uses two feature encoding vectors with features aggregated from the width and height of the image/layer features. Therefore, spatial information can be incorporated into the network for prediction and classification purposes. Additionally, the Coordinate Attention module is built in a modularized way that supports incorporating itself into YOLOv7 models easily via inserting simple codes and changing configuration files.

\section{Data Collection and Benchmark Datasets}
Recent developments in smartphones, dashcams, and drones enable automated roadway damage data collection. The smartphone and dashcam approach is easier to set up and implement and more efficient than the manual approach to scanning and finding roadway damages. However, it still demands human drivers to drive through the roads and record videos. Conversely, we can utilize drones to reduce the human workload. 

Labeling road damage data manually on the collected images to create training datasets would be laborious. Therefore, we approach existing datasets as benchmarks for training deep learning models to automatically detect road damage from the collected videos. There are some datasets available. However, the one provided in the IEEE 2020 Big Data Challenge Cup by Sekilab\footnote{\url{https://rdd2020.sekilab.global/}} is considered practical.

\begin{figure}[!htb]
    \centering
    \includegraphics[width=\linewidth]{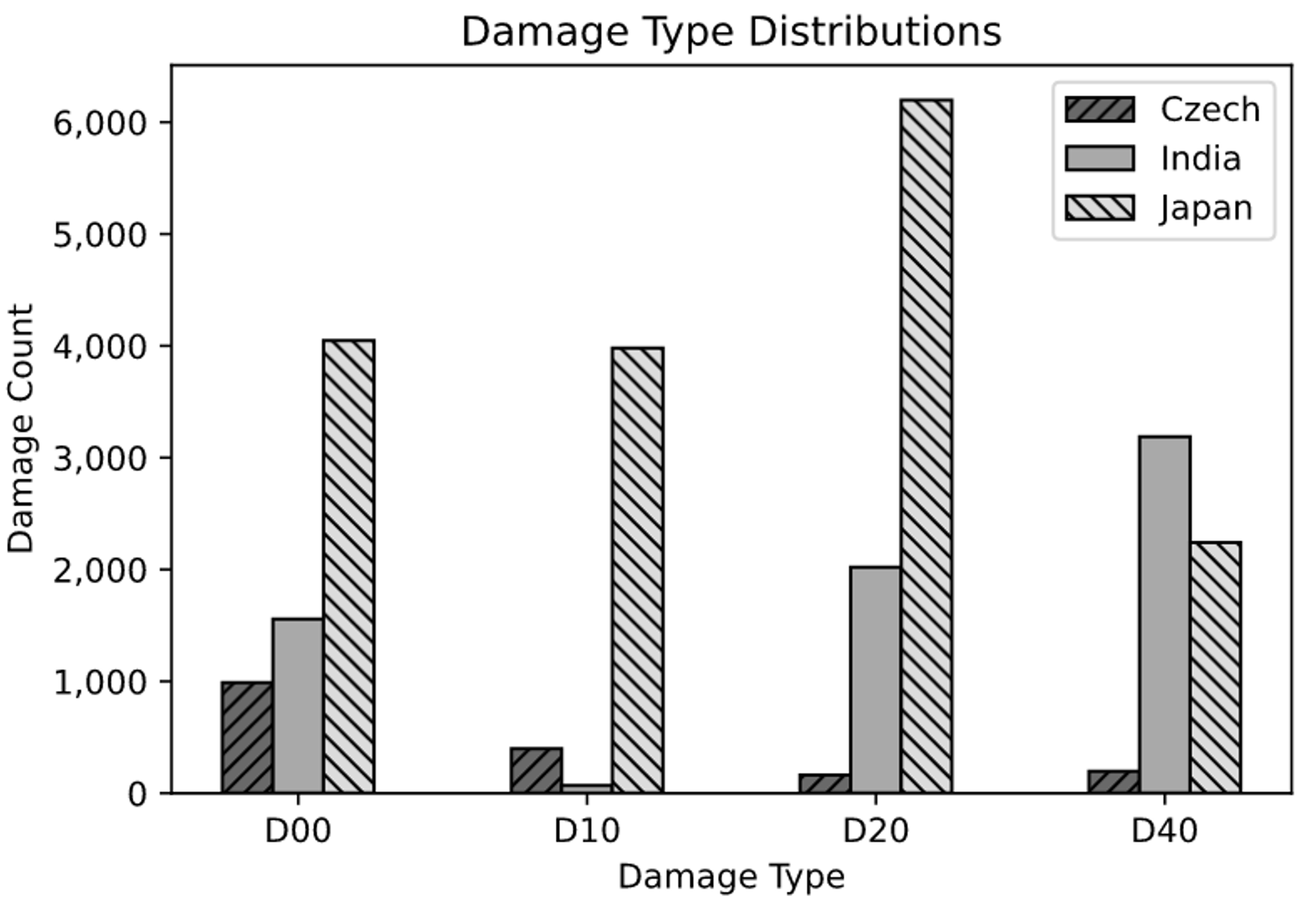}
    \caption{Distribution of damage types in the current benchmark dataset. D00, D10, D20, and D40 are longitudinal cracks, traverse cracks, alligator cracks, and potholes, respectively.}
    \label{fig:rdd2020}
\end{figure}

This benchmark dataset consists of one training set (\textit{train}) and two test sets (\textit{test1} and \textit{test2}). The training set contains 21,041 images (2,829, 7,706, and 10,506 for Czech, India, and Japan, respectively). The two test sets contain 2,631 and 2,664 images, correspondingly. The training set has 34,702 ground-truth labels (bounding boxes and damage types). There are four damage types considered: longitudinal cracks (D00), traverse cracks (D10), alligator cracks (D20), and potholes (D40). Specifically, Figure \ref{fig:rdd2020} shows the damage type distributions (of the four corresponding damage types) over the three countries.

Our experiments with this benchmark dataset showed that models trained on the dataset of one country do not work well in other countries because they have different road types. Therefore, if we trained deep learning models on this dataset, it would not scale well to the roads in the US. Consequently, we collected a separate dataset for the USA. Specifically, we utilized Google Street View API\footnote{\url{https://developers.google.com/maps/documentation/streetview/overview}} to download images from Google Street View. 

Google Street View is an excellent resource with a vast number of images of roads all over the world. We focused only on states in the US in this current project. However, we can theoretically go to any city in the world that Google Street View supports to collect data on road damage. Furthermore, Google also updates its images pretty frequently. Specifically, most of the collected images in this dataset were captured in 2020 or later. Also, the highest downloadable image resolution is 640 by 640, which is suitable for training deep learning models.

\begin{figure}[!htb]
    \centering
    \includegraphics[width=\linewidth]{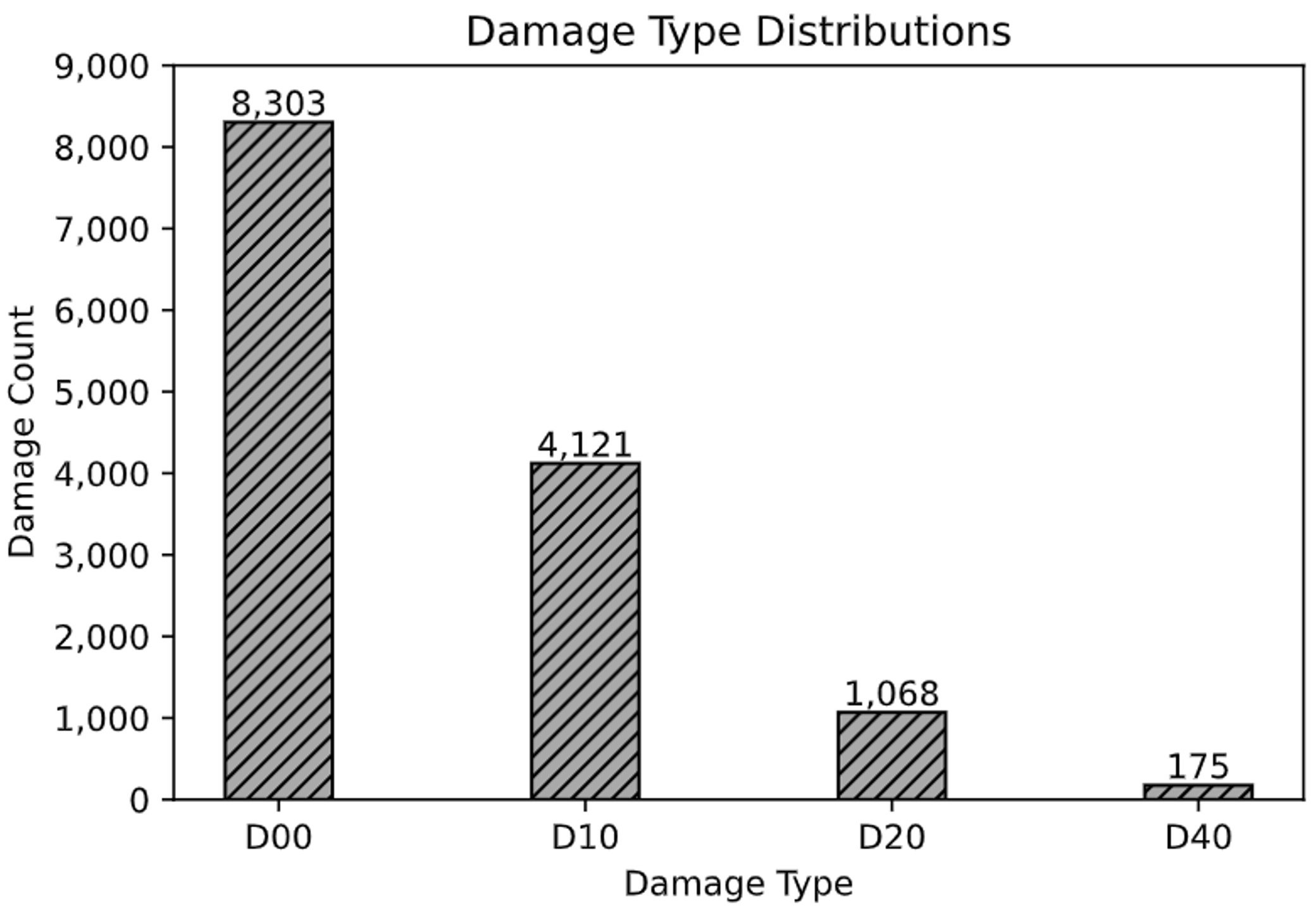}
    \caption{Distribution of damage types in the newly collected set of images from the United States using Google Street View. There are four types of damages considered: longitudinal (D00), traverse (D10), alligator (D20), and pothole (D40) damages, respectively.}
    \label{fig:usdamagedistributions}
\end{figure}

In this project, we collected 6,005 images from Google Street View. As shown in Figure \ref{fig:usdamagedistributions}, there are 8,303; 4,121; 1,068; and 175 damages for longitudinal (D00), traverse (D10), alligator (D20), and pothole (D40) damages, correspondingly. With this dataset, we were awarded the silver prize for data contributor in the Crowdsensing-based Road Damage Detection Challenge (CRDDC2022), IEEE BigData 2022\footnote{\url{https://crddc2022.sekilab.global/}}.

In this same competition, other institutions/organizations also contributed their images. These are valuable sources for increasing labeled images for training deep learning models. Therefore, this project utilizes this as a benchmark dataset to explore road damage detection and classification tasks using YOLOv7. Specifically, this dataset is available online~\cite{rdd2022}. Table~\ref{tab:traintestimages} shows the corresponding numbers of images for training and testing in this dataset.

\begin{table}[htbp]
\caption{The numbers of (roughly) train and test images from six different countries in the Crowdsensing-based Road Damage Detection Challenge (CRDDC2022).}
\begin{center}
\begin{tabular}{|l|r|r|}

\hline
\textbf{Country} & \textbf{Train images} & \textbf{Test images} \\
\hline
India & 9,000 & 1,000\\
\hline
Japan & 10,000 & 1,000 \\
\hline
Czech Republic & 2,000 & 1,000 \\
\hline
Norway & 9,000 & 1,000 \\
\hline
United States$^{\mathrm{a}}$ & 5,000 & 1,000 \\
\hline
China & 3,500 & 1,000 \\
\hline
\multicolumn{3}{l}{$^{\mathrm{a}}$This is our contribution.}
\end{tabular}
\label{tab:traintestimages}
\end{center}
\end{table}

\section{Data Processing}
\subsection{Image sizes}
After training an initial model, it turned out that the prediction and classification accuracy were low. One of the reasons is that we need to reduce the training image size to $640\times640$ (the standard training image size for YOLOv7) for all countries because of the limitation of our device's GPUs. Therefore, we would like to have a deeper look at image sizes from countries and how they are related to prediction accuracy. As shown in Table~\ref{tab:imagesizes}, it turns out that Japan and Norway have different image sizes within themselves. This might not be a problem for YOLOv7 because YOLOv7 automatically transforms images and corresponding labels into standard sizes before training. However, it might be a problem for those who would like to convert the YOLOv7 annotation format (ratio-based, floating numbers from 0 to 1) to absolute pixel locations if assuming the same image size for all images. Notably, United\_State is the only folder that has YOLOv7 standard training image size.

\begin{table}[htbp]
\caption{Image sizes for different image folders in the Crowdsensing-based Road Damage Detection Challenge (CRDDC2022).}
\begin{center}
\begin{tabular}{|l|r|r|}

\hline
\textbf{Folder} & \textbf{Image width(s)} & \textbf{Image height(s)} \\
\hline
China\_Drone & {512} & {512}\\
\hline
China\_MotorBike & {512} & {512}\\
\hline
Czech & {600} & {600}\\
\hline
India & {720} & {720}\\
\hline
Japan &{600, 1024, 1080, 540} & {600, 1024, 1080, 540}\\
\hline
Norway & {4040, 3650, 3643} & {2041, 2035, 2044}\\
\hline
United\_States$^{\mathrm{a}}$ & {640} & {640} \\
\hline
\multicolumn{3}{l}{$^{\mathrm{a}}$YOLOv7 standard training image sizes.}
\end{tabular}
\label{tab:imagesizes}
\end{center}
\end{table}

Our initially trained models also indicate that the prediction and classification accuracy for the Norway folder is relatively low compared to data from other folders. The starting point of our investigation into this is the image sizes. A deeper investigation into Norway's set of images indicates several reasons for this low accuracy. These reasons include having different image sizes, too large images, and many images in this folder do not have labels (thus do not help training). As discussed, having different image sizes may not be a problem; however, having too large images is. Two main issues include RAM consumption and label resizing. Specifically, large RAM consumption impacts training and inference time (our current standard 16GB GPU RAM is not able to perform training on these image sizes). Thus, these images are scaled to standard image sizes ($640\times640$) before training. Additionally, there are also many small labels in these large images. Thus, scaling to smaller image sizes makes several road damage labels become too small (even smaller than 1 pixel in the scaled size). Road damage annotation with too small (or invalid) sizes reduces the models' performances.

\begin{figure}[!htb]
    \centering
    \includegraphics[width=\linewidth]{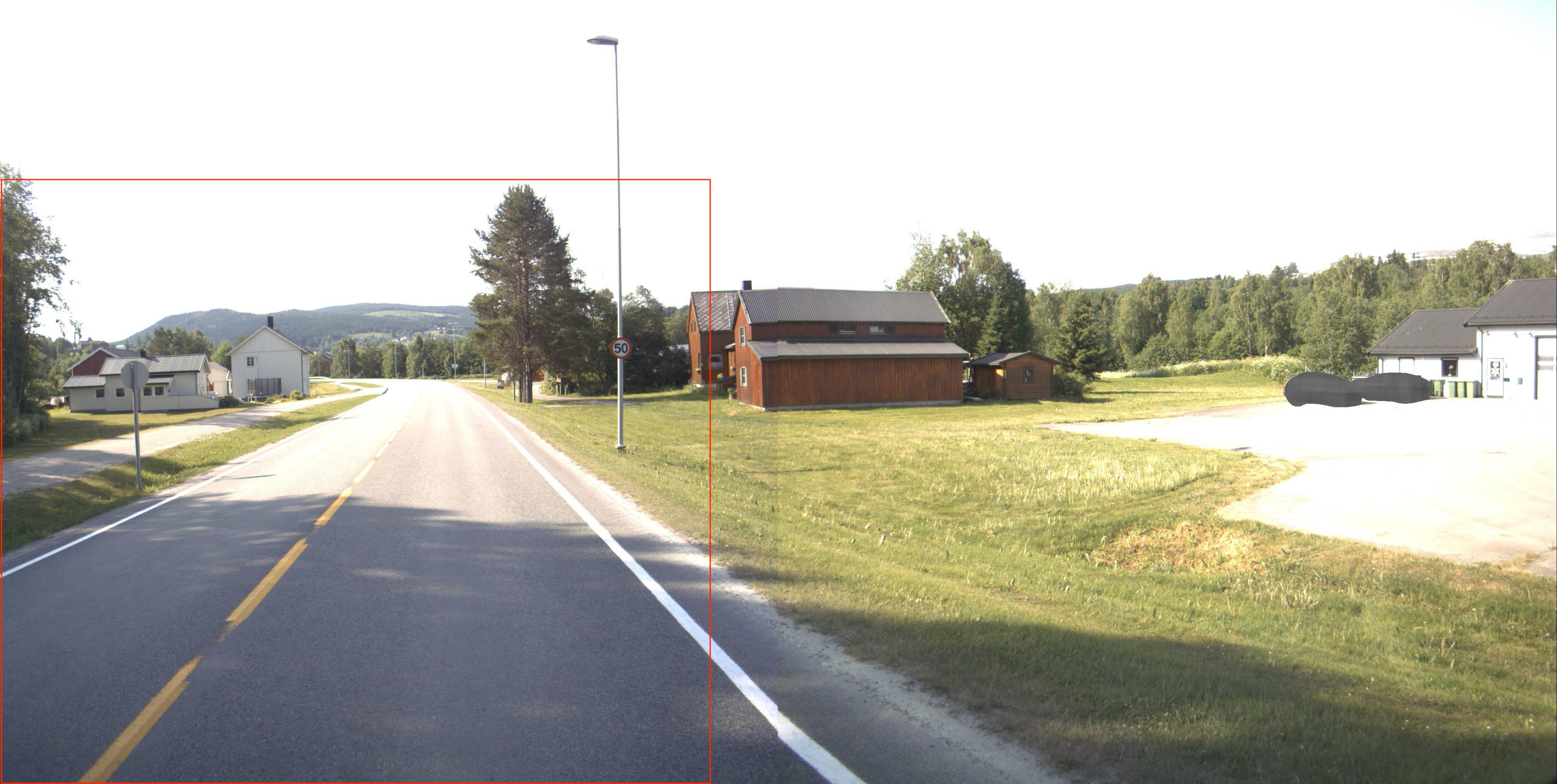}
    \caption{A typical image from the Norway folder (Norway\_000003.jpg). It has a large image size and seems to be a patch of two images (tiling from left to right). Most of the road section (and damages) are at the lower-left corner of the image (the $1824\times1824$ pixels surrounded by the red rectangle).}
    \label{fig:norwayimage}
\end{figure}

A further look at the pictures in this folder reveals some interesting details. Some pictures seem to be patches of two images together (one on the left and another one on the right). As shown in Figure~\ref{fig:norwayimage} for \textit{Norway\_000003.jpg} in this folder, most pictures have the roads located at the lower-left corner. Therefore, we decided to crop these pictures and take only $1824\times1824$ pixels at the lower-left corner of every picture from the Norway folder. The reason for selecting 1824 pixels is because it is about half of the width of the original picture where most of the road is located, and YOLOv7 requires a training image of size as multiple of 32. Additionally, we also crop or remove the annotations for damages outside of this region. The number of cropped/removed images is insignificant (537 out of 10,692 road damages), meaning that the selected region is appropriate. Lastly, we name the resulting folder as \textit{Norway1} in this dataset.
%\subsection{Notation conversion}
\subsection{Annotation cleaning and train/validation division}
\begin{figure}[!htb]
    \centering
    \includegraphics[width=\linewidth]{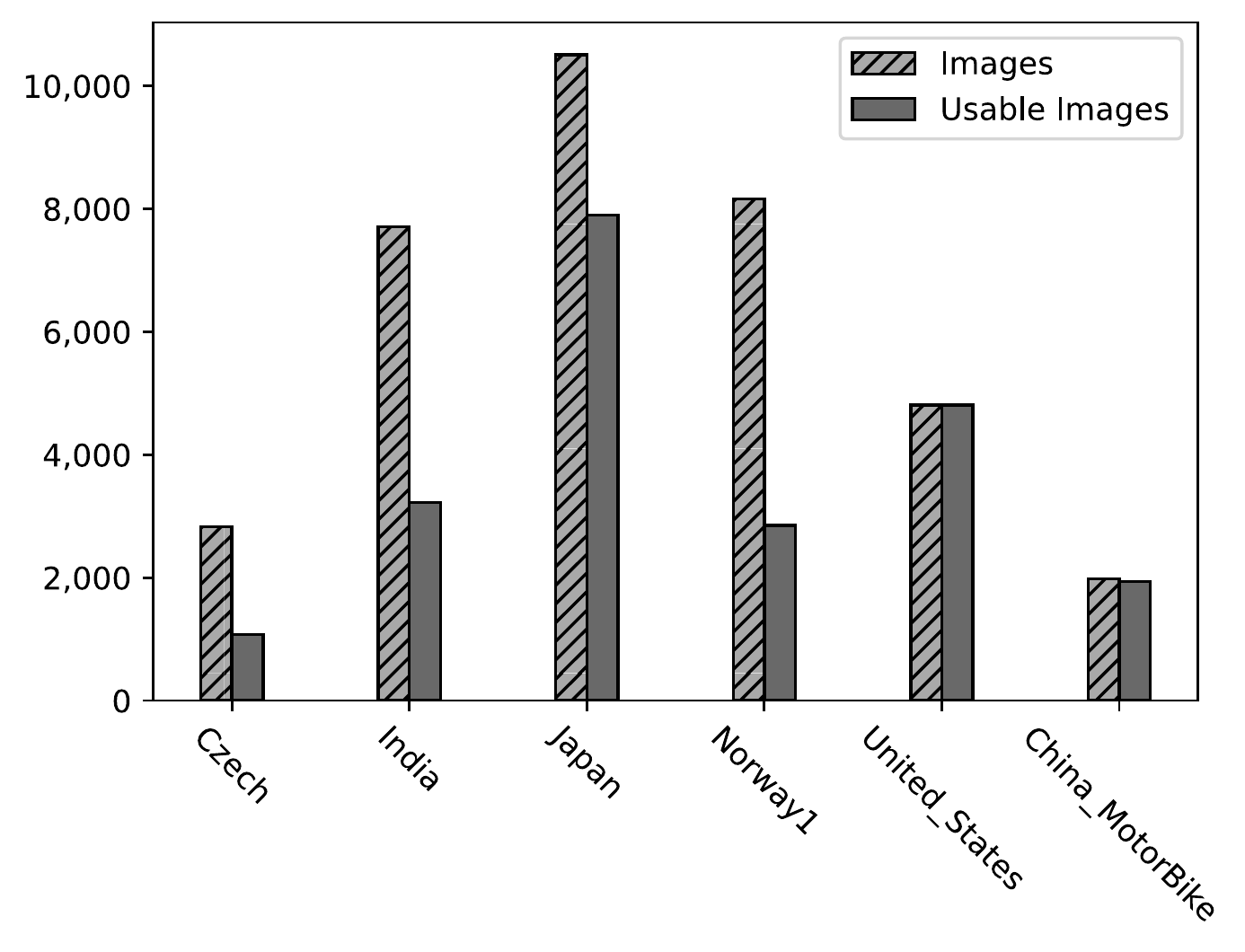}
    \caption{Distribution of numbers of images and images usable for training (images with damage annotations) in the six folders (folders that have test data).}
    \label{fig:usableimages}
\end{figure}

In the data folders of this dataset, there are images that do not have annotations (e.g., many of them in the Norway folder) and images that have annotations that are not one of the four types considered in this project (D00, D10, D20, D40). Figure~\ref{fig:usableimages} depicts the number of images in the folders with a test set and the corresponding number of images usable for training (those that do not have annotations cannot be utilized for training). Observably, Japan, Norway, and India are the top three with the high number of images, while Japan, United\_States, and India are the top three with the highest number of usable images.

\begin{figure}[!htb]
    \centering
    \includegraphics[width=\linewidth]{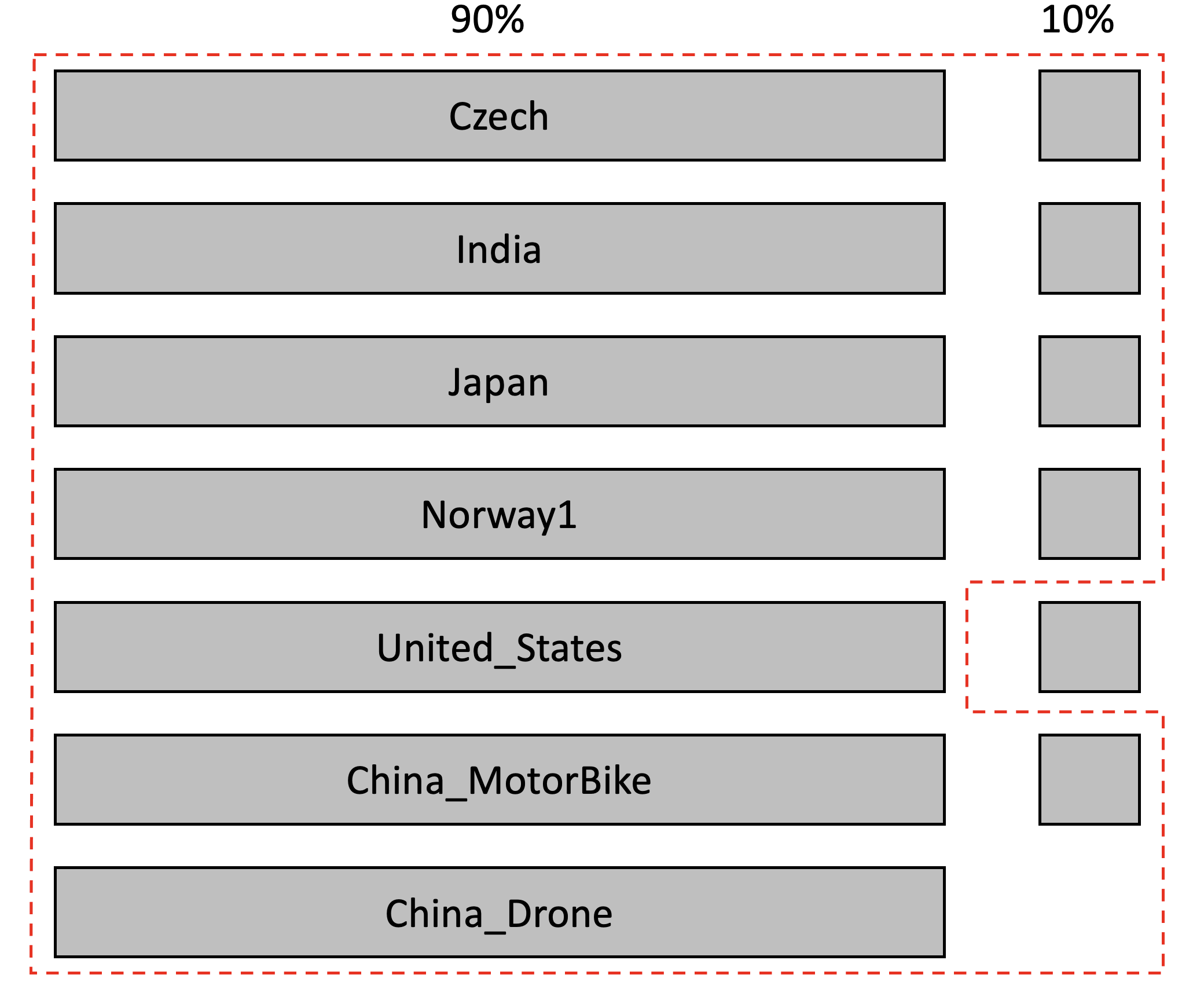}
    \caption{Data division strategy. For instance, the best models for United\_States are selected by 10\% of its images (as validation set), and all other images are placed into the training set (red dashed box indicates the train data for training models for United\_States).}
    \label{fig:trainvalsplit}
\end{figure}

Out of the seven data folders, China\_Drone does not have a test set. Therefore, we only create models for those that have a test dataset. Figure~\ref{fig:trainvalsplit} depicts the strategy used to split images into the training set (\textit{train}) and validation set (\textit{val}) used in this project. Specifically, China\_Drone does not have a test set; thus, its images are all placed into the training set. For each of the other folders, images are divided into 90\% and 10\%. When training models for one folder, 10\% of that folder is placed into the validation set, and all other images are placed into the training set.

For instance, for the United\_States case, we place 10\% of its images into the validation set. Then we place the rest 90\% of its images together with images from all other countries (including China\_Drone) into the training set. The reason for this is that we need to use genuine United\_States data in the validation set to select the best model for United\_States. However, the training set can include a mix-up of data from different sources to increase the amount of training data besides augmentation. Similarly, we perform the same splitting strategy for other folders.

\section{Experiments}
\subsection{Evaluation Metrics}
Different experiments result in different models. Thus, we need to have a robust metric to select the best models out of all experiments. There are two common evaluation metrics used in this area. The first one is the Average Precision (mAP) calculated at IoU (Intersection over Overlapping) threshold of 0.5 (mAP@0.5). The second one is the F1 score. 

% to continue check Grammarly from here.

The mAP is a good measurement when we need to ensure the model is stable across different confidence thresholds (robust) while the F1 score is computed for a specific confidence threshold. The common practice is to use mAP@0.5 on the validation set to select the best model and use the F1 score to report the model performance on the test dataset. This project also follows this common practice (using mAP@0.5 to select the best models and report F1 scores on the test sets).

% may want to add some details here: https://stackoverflow.com/questions/62973155/high-map50-with-low-precision-and-recall-what-does-it-mean-and-what-metric-sho

\subsection{Train image augmentations}
\begin{figure}[!htb]
    \centering
    \includegraphics[width=\linewidth]{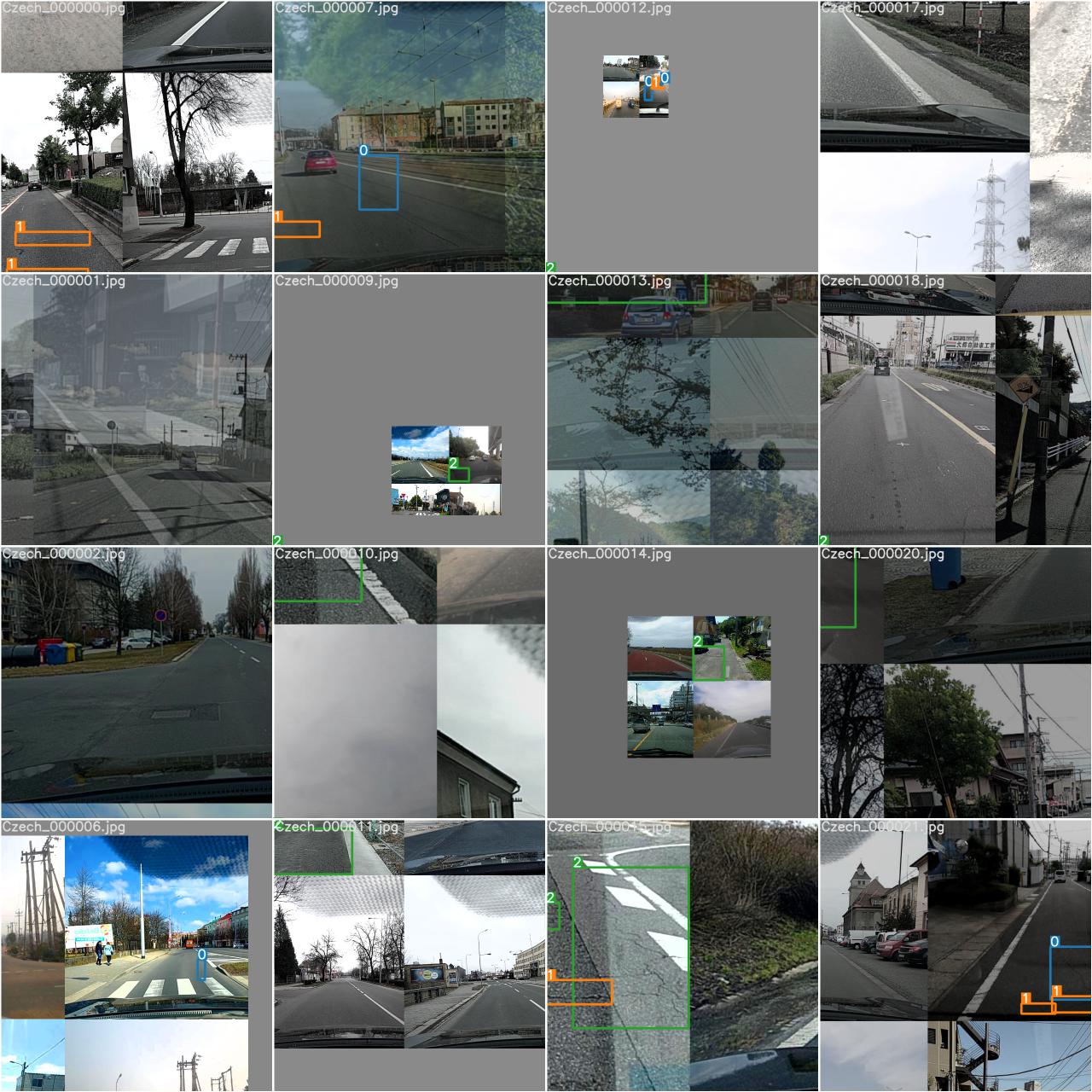}
    \caption{One example of augmented images using YOLOv7's default parameters.}
    \label{fig:yoloaugmentations}
\end{figure}

Image augmentation is a great technique to improve model robustness and accuracy. YOLOv7 uses the default image augmentations as listed in Table~
\ref{tab:yoloaugmentations}. However, Figure~\ref{fig:yoloaugmentations} indicates that there are several obvious issues with these default parameters. Specifically, the scale parameter is too large, which makes some pictures become too small (thus, the damages). The mosaic, mixup, and paste\_in options are high, making the augmented images unrealistic. 

\begin{table}[htbp]
\caption{YOLOv7\protect\footnote{https://github.com/WongKinYiu/yolov7} default image augmentation parameters.}
\begin{center}
\begin{tabular}{|l|r|l|}
\hline
\textbf{Parameter} & \textbf{Value} & \textbf{Descriptions} \\
\hline
hsv\_h & 0.015  &   HSV-Hue augmentation (fraction) \\
\hline
hsv\_s & 0.7  &   HSV-Saturation augmentation (fraction) \\
\hline
hsv\_v &  0.4  &   HSV-Value augmentation (fraction) \\
\hline
degrees &  0.0  &   rotation (+/- deg) \\
\hline
translate &  0.2  &   translation (+/- fraction) \\
\hline
scale &  0.9  &   scale (+/- gain) \\
\hline
shear&  0.0  &   shear (+/- deg) \\
\hline
perspective &  0.0  &   perspective (+/- fraction)\\
\hline
flipud &   0.0  &   flip up-down (probability) \\
\hline
fliplr &  0.5  &   flip left-right (probability) \\
\hline
mosaic &  1.0  &   mosaic (probability) \\
\hline
mixup &  0.15  &   mixup (probability) \\
\hline
copy\_paste &  0.0  &   copy paste (probability) \\
\hline
paste\_in &  0.15  &   copy paste (probability)\\
\hline
loss\_ota &  1 &  use ComputeLossOTA \\
\hline
\end{tabular}
\label{tab:yoloaugmentations}
\end{center}
\end{table}

\begin{figure}[!htb]
    \centering
    \includegraphics[width=\linewidth]{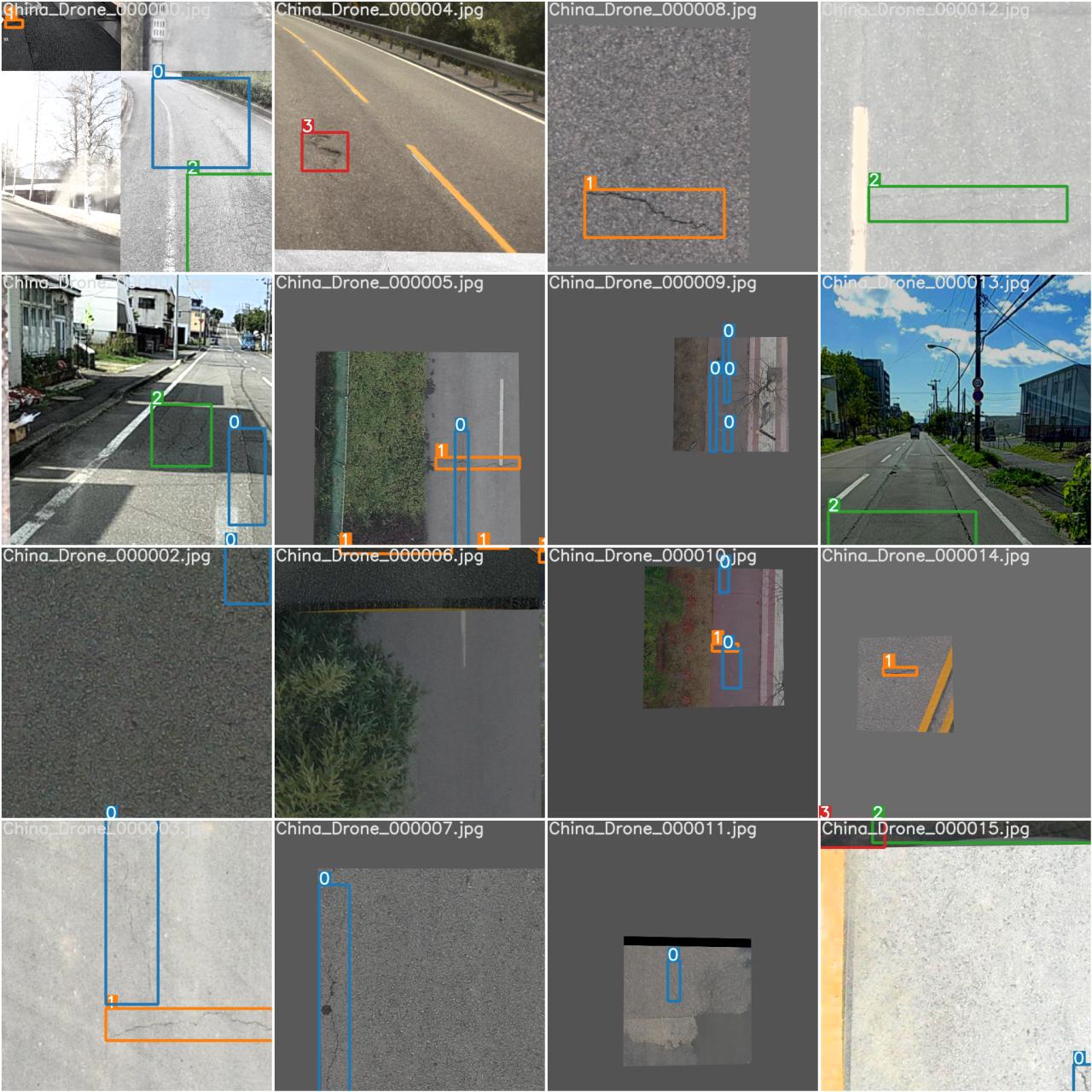}
    \caption{One example of augmented images using proposed augmentation parameters.}
    \label{fig:imageaugmentations}
\end{figure}

Table~\ref{tab:imageaugmentations} shows the final list of image augmentation parameters used in this project (other default parameters remain the same). Specifically, scale, mosaic, mixup, and paste\_in are slightly reduced to avoid unwanted, unrealistic effects. Additionally, as discussed previously, road damages collected using cameras placed on car dashboards have some perspective. Thus, we utilized shear (0.01) and perspective (0.0001) in the image augmentation parameters. Figure~\ref{fig:imageaugmentations} shows an example of images in a training epoch generated with this set of hyperparameters. Observably, these augmented images are more realistic than those in Figure~\ref{fig:yoloaugmentations}.

\begin{table}[htbp]
\caption{Experimented image augmentation parameters (other parameters remain the same).}
\begin{center}
\begin{tabular}{|l|r|l|}
\hline
\textbf{Parameter} & \textbf{Value} & \textbf{Descriptions} \\
\hline
scale &  0.7  &   scale (+/- gain) \\
\hline
shear&  0.01  &   shear (+/- deg) \\
\hline
perspective &  0.0001  &   perspective (+/- fraction) \\
\hline
mosaic &  0.5  &   mosaic (probability) \\
\hline
mixup &  0.1  &   mixup (probability) \\
\hline
paste\_in &  0.05  &   copy paste (probability)\\
\hline
\end{tabular}
\label{tab:imageaugmentations}
\end{center}
\end{table}

\subsection{Coordinate Attentions}
Given an input X with height H, width W, and C channels, Coordinate Attention~\cite{hou2021coordinate} uses two pooling kernels of size (H, 1) and (1, W) to extract features from spatial directions accordingly. Specifically, the output of a channel $c$ at width w and height $h$ can be computed using the following equations:
\begin{equation}
    z^h_c = \frac{1}{W} \sum_{i=0}^{W-1} x_c(h, i)
\end{equation}
\begin{equation}
    z^w_c = \frac{1}{H} \sum_{j=0}^{H-1} x_c(j, w)
\end{equation}
where $x_c(x, y)$ is the value of channel $c$ at location $(x, y)$.

\begin{equation}
    f = \delta \bigl( F1([z^h, z^w]) \bigr)
\end{equation}
where $[z^h, z^w]$ is the concatenation of two computed spatial features (a pair of two 1D vectors of sizes $W$ and $H$ for each channel) to make C one 1D vector of size $W + H$. Additionally, $F1$ is a convolutional transformation with a kernel size of $1\times1$ and $C/r$ number of filters ($r$ is called the reduction ratio). Finally, $\delta$ is a nonlinear activation function. Consequently, the resulting feature is of size $C/r\times(W + H)$.

The resulting feature $f$ is then split into two separate parts (according to $W$ and $H$) to get $f^h$ of size $C/r \times H$ and $f^w$ of size $C/r \times W$. These produced features are then transformed by two transformation functions ($F_w$ and $F_h$). They are two different convolutional layers with $1\times1$ kernel size and the number of filters as $C$ (same size as the input channels). These transformations generate two feature vectors:
\begin{equation}
    g^h = \sigma\bigl(F_h(f^h)\bigr)
\end{equation}

\begin{equation}
    g^w = \sigma\bigl(F_w(f^w)\bigr)
\end{equation}
where $\sigma$ is the sigmoid function that helps to convert the values into the range of 0 to 1 (as the weights), the output $Y$ then can be computed as:
\begin{equation}
    y_c(i, j) = x_c(i, j) \times g^h_c(i) \times g^w_c(j)
\end{equation}
Notably, $Y$ has the same shape as the input X, but it is now weighted to incorporate spatial coordinates.

The selected YOLOv7 model for this project has 50 layers in the backbone. We first trained models with this default configuration for each of the data folders. Next, we added three coordinate attentions to the layers right before sending to the three last \textit{RepConv} (Represented Convolution~\cite{ding2021repvgg}) layers. These layers are selected because their outputs are fed directly to the last detect layer (\textit{IDetect}).

These three additional Coordinate Attention layers significantly increase the model performance without noticeable inference time changes. Therefore, we decided to add three other coordinate attention layers to the backbone network. They are placed after layers 24, 37, and 50 correspondingly. These layers are selected because their outputs are what passed from YOLOv7's backbone to YOLOv7's head (thus, we applied coordinate attention to these outputs before passing them to the head). These other three additional coordinate attentions also improve the accuracy significantly.

\subsection{Label smoothing}
For various reasons, the annotated labels are not 100\% accurate. For instance, many traverse cracks with different perspectives or a slight camera rotation may be mislabeled as longitudinal cracks and vice versa. Additionally, alligator cracks are often confused with longitudinal and traverse cracks. Therefore, label smoothing~\cite{szegedy2016rethinking} technique helps improve performance in this case. Precisely, classification loss from classification heads (classifying road damages into one of the four considering crack types) is calculated using cross-entropy using the following formula:
\begin{equation}
\label{eq:crossentropy}
    \mathit{loss} = \sum_{c=1}^{C} \bigl (- y_c \log (\hat y_c) - (1-y_c) \log(1 - \hat y_c) \bigr)
\end{equation}
where $C$ is the number of classes, $\hat y_c$ is the predicted probability of class label $c$ and $y_c$ is calculated as:
\begin{equation}
\label{eq:hardlabel}
    y_c = 
    \begin{cases}
        1~\text{if the item is of class c} \\
        0~\text{otherwise}
    \end{cases}
\end{equation}
This way of setting the value for $y_c$ is called hard label (100\% sure that the label is of class $c$). However, label smoothing allows having a smoothing constant $\epsilon$ that softens this confidence:
\begin{equation}
\label{eq:labelsmoothing}
    y_c = 
    \begin{cases}
        1-\epsilon + \epsilon/C~\text{if the item is of class c} \\
        \epsilon/C~\text{otherwise}
    \end{cases}
\end{equation}
The common value for $\epsilon$ is 0.1, which is what we experimented with in this project. Notably, the values of all $y_c$ over all classes still add up to one ($\sum_c^Cy_c = 1$).

\subsection{Additional accuracy fine-tuning techniques}
We build three models for each dataset folder with a test set. One model is the default YOLOv7 configuration with the modified image augmentation options, one model is the configuration with three additional coordinate attention layers in YOLOv7's head, and one model is the configuration similar to the second one and three more Coordinate Attention layers in the YOLOv7's backbone. We use the ensemble method to combine the results from the three best models for each folder and improve the accuracy. Furthermore, Japan and India have similar sets of images. Therefore, we use the ensemble of the best models of both countries to perform inference on the test sets of these two countries. Additionally, we also applied test time augmentation techniques (using \textit{--augment} option in YOLOv7's inference command).

These fine-tuning techniques improve prediction accuracy with a slight trade-off of the inference time compared to the standard YOLOv7 inference time. However, this trade-off is insignificant because standard YOLOv7 with standard configuration (used in this project) is reasonably fast ($\approx40$ 114 frames per second). These techniques are highly recommended and help boost prediction accuracy if time requirements are insignificant.

\section{Results and discussions}
The mAP@0.5 measurements are used to select the best models at the training time based on the validation data. However, the models' performances on the test sets are evaluated using the F1 score from the organizer. The prediction results on the test sets produced by experimented models and their ensembles are uploaded to the dataset's publisher. They divide the evaluation submissions into five leaderboards. One board per folder for India, Japan, Norway, United States, and one folder for the test images from all six countries called "Overall" (including also Czech and China\_Drone). The average of all six boards is used to rank submissions to these leader boards.

\begin{table}[htbp]
\caption{F1 scores on the leader boards\protect\footnote{https://crddc2022.sekilab.global/leaderboard/}}
\begin{center}
\begin{tabular}{|c|c|c|c|c|c|}
\hline
\textbf{Average} & \textbf{Overall} & \textbf{India}  & \textbf{Japan}  & \textbf{Norway}  & \textbf{United States} \\
\hline
0.663 & 0.741 &  0.516 &  0.735 & 0.504 & 0.817 \\
\hline
\end{tabular}
\label{tab:leaderboards}
\end{center}
\end{table}

\begin{figure}[!htb]
    \centering
    \includegraphics[width=\linewidth]{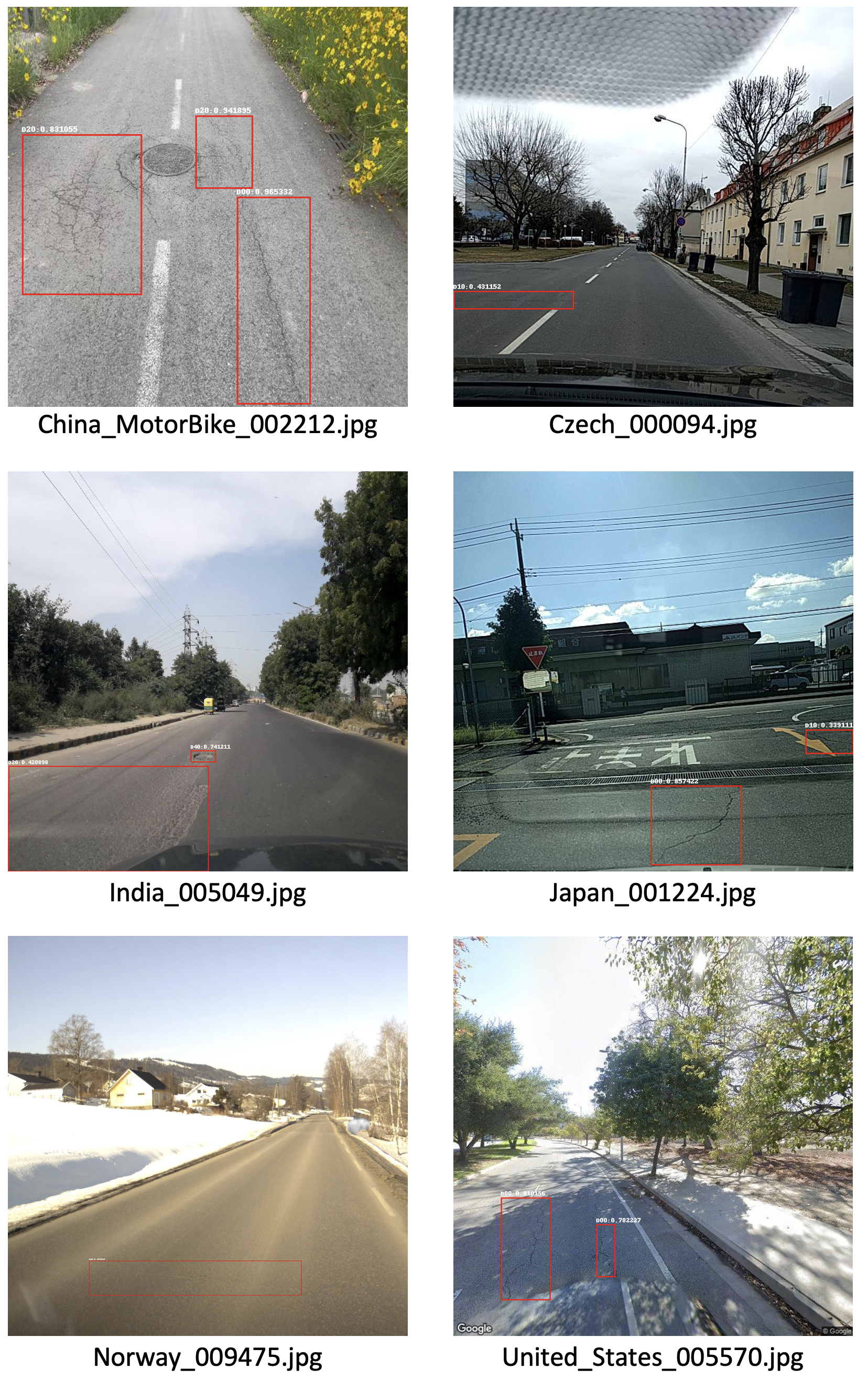}
    \caption{Examples of damage detection from road images using the approach proposed in this paper.}
    \label{fig:detectionexamples}
\end{figure}

Table~\ref{tab:leaderboards} summarizes the F1 scores on these boards for our proposed models. Notably, Japan and United States boards have the highest accuracy because they have the most usable training images (as in Figure~\ref{fig:usableimages}). The United States has a little higher accuracy compared to Japan, probably because there are more longitudinal damages in this folder (as shown in Figures~\ref{fig:usdamagedistributions} and~\ref{fig:rdd2020}. The longitudinal is considered easier to classify. Similarly, India and Norway have similar F1 scores, and they are lower because they have lower and similar numbers of usable images. Figure~\ref{fig:detectionexamples} shows examples of the damages detected by our proposed models for the test images in the six test sets. Lastly, the average accuracy of all boards of 0.663 ranked our approach the $3^{rd}$ place among 51 teams that participated in the Crowdsensing-based Road Damage Detection Challenge (CRDDC2022), IEEE BigData 2022 Cup\footnote{https://crddc2022.sekilab.global/leaderboard/}. 

The prediction and classification accuracies can be improved in several ways. First, collecting road data from Google Street View using our proposed approach is relatively easy. We will collect and label more data to train more robust models. Additionally, if the computation power (e.g., GPUs) is available, we would like to try other YOLOv7 configurations\footnote{https://github.com/WongKinYiu/yolov7} (YOLOv7-X, YOLOv7-W6, YOLOv7-E6, YOLOv7-D6, YOLOv7-E6E). These models were proved to have better accuracy on the MS COCO (Microsoft Common Objects in Context) dataset, but we could not experiment with them due to computation limitations. Furthermore, if the inference time is not a strong requirement, many more models can be built to create an ensemble with higher accuracy.

Finally, the experiments' source codes, data files, and configuration files are available on this project's GitHub page: \url{https://github.com/mdptlab/roaddamagedetector2022}.

\section{Conclusion}
This work proposes to use Google Street View to collect and label road damages. This data collection approach is efficient because Google Street View has a huge amount of images of roads all over the world. These images are also frequently updated, which helps to get the latest road conditions. Additionally, this work explores different state-of-the-art object detection methods and their applicability for road damage detection and classification tasks. We experiment with YOLOv7 with different configurations, incorporating coordinate attentions and label smoothing techniques. We also utilized fine-tuning techniques such as image augmentations and an ensemble method to improve accuracy. 

These proposed approaches are applied to the dataset from the Crowdsensing-based Road Damage Detection Challenge (CRDDC2022), IEEE BigData 2022, and yield state-of-the-art results on the test images. Specifically, the proposed approach has the F1 scores of 81.7\% on the road damage data collected from the United States using Google Street View and 74.1\% on all test images of this dataset. In the future, we suggest using Google Street View and collecting more road damage data to train more robust models with higher accuracy. Additionally, if computation resources allow, we would like to explore more complicated (heavier) YOLOv7 backbone networks available on YOLOv7's GitHub page. These models were proved to have higher accuracy on MS COCO (Microsoft Common Objects in Context) dataset.

\bibliographystyle{./IEEEtran}
\bibliography{IEEEabrv,IEEEfull}

\end{document}